\documentclass[letterpaper, 10 pt, conference]{ieeeconf}  % Comment this line out if you need a4paper

\IEEEoverridecommandlockouts                              % This command is only needed if 
\usepackage{graphics} % for pdf, bitmapped graphics files
\usepackage[pdftex]{graphicx}
\graphicspath{{./Figures/,./}}
\DeclareGraphicsExtensions{.pdf,.jpeg,.png}
\usepackage{epsfig} % for postscript graphics files
\usepackage{amsmath} % assumes amsmath package installed
\usepackage{amssymb}  % assumes amsmath package installed
\usepackage{bbold}
\usepackage[export]{adjustbox}
\usepackage[]{subfig}
\usepackage{lipsum}
\usepackage{tabularx,booktabs}
\usepackage{todonotes}
\usepackage{textcomp}
\usepackage{float}
\usepackage{cuted}
\usepackage{todonotes}
\newcommand{\R}{\mathbb{R}}

\usepackage{cuted}
\usepackage{tikz}
\usepackage{float}
\usepackage{xfrac}

\DeclareMathOperator{\proj}{\textbf{proj}}

\usepackage[font=footnotesize]{caption}

\newtheorem{remark}{Remark}
\newtheorem{assumption}{Assumption}
\usepackage{color}
\usepackage{microtype}
\renewcommand{\baselinestretch}{0.98}

\makeatletter
\newcommand\mypicture[2][\textwidth]{%
  \setbox0\hbox{\includegraphics[width=\textwidth]{#2}}
  \ifdim\ht0>\dimexpr\pagegoal-\pagetotal\relax
    \@latex@warning{This picture might be oddly placed}\fi
  \begin{strip}
    \includegraphics[width=#1]{#2}
  \end{strip}
}
\makeatother

\title{\LARGE \bf
Identification of Compliant Contact Parameters and Admittance Force Modulation on a Non-stationary Compliant Surface 
}
\author{\large Lasitha Wijayarathne, \textit{IEEE Student Member} and  Frank L. Hammond III, \textit{IEEE Member}% <-this % stops a space
\thanks{Lasitha Wijayarathne and Frank L. Hammond III are with the Woodruff School of Mechanical Engineering and the Coulter Department of Biomedical Engineering at the Georgia Institute of Technology. {\tt\small frank.hammond@me.gatech.edu.}}%
}
\begin{document}
\maketitle
\thispagestyle{empty}
\pagestyle{empty}
%%%%%%%%%%%%%%%%%%%%%%%%%%%%%%%%%%%%%%%%%%%%%%%%%%%%%%%%%%%%%%%%%%%%%%%%%%%%%%%%
\begin{abstract}
Although autonomous control of robotic manipulators has been studied for several decades, they are not commonly used in safety-critical applications due to lack of safety and performance guarantees - many of them concerning the modulation of interaction forces. This paper presents a mechanical probing strategy for estimating the environmental impedance parameters of compliant environments, independent a manipulator's controller design, and configuration. The parameter estimates are used in a position-based adaptive force controller to enable control of interaction forces in compliant, stationary, and non-stationary environments. This approach is targeted for applications where the workspace is constrained and non-stationary, and where force control is critical to task success. These applications include surgical tasks involving manipulation of compliant, delicate, moving tissues. Results show fast parameter estimation and successful force modulation that compensates for motion.  
\end{abstract}

%%%%%%%%%%%%%%%%%%%%%%%%%%%%%%%%%%%%%%%%%%%%%%%%%%%%%%%%%%%%%%%%%%%%%%%%%%%%%%%% 
\section{Introduction}
The control of the contact forces between a robot and its environment is essential to a variety of safety-critical applications. Some examples include force modulation on compliant environments, such as interactions in a surgical setting, micro-assembly, or biological tissue manipulation. In the literature, a study on force control dates back about four decades, and various control methods have been intensively studied and applied\cite{SicilianoB.SciaviccoL.VillaniL.OrioloRobotAnalysis}. \par Despite that, applications of force control schemes in safety-critical domains are not widely utilized due to the complexity involved in the implementation control scheme, incompatibility with industrial controllers, spatial constraints, or inability to adapt to temporal or spatial variations in impedance parameters. To mitigate the above problems, model-based and data-driven techniques have been studied to improve force control; however, they lack guarantees on safety and force tracking performance\cite{Lee2015LearningDemonstrations} as desired in uncertain environments.  For successful integration of force control techniques for manipulators working in uncertain environments, usage of simplistic robust models, adaptable control schemes, and effective feedback control schemes are essential to developing robust schemes with desired performance. Potential applications of this could be any industrial setting where force control in a compliant environment is needed. 
 
\begin{figure}[t!] 
\centering
\includegraphics[width=.9\linewidth]{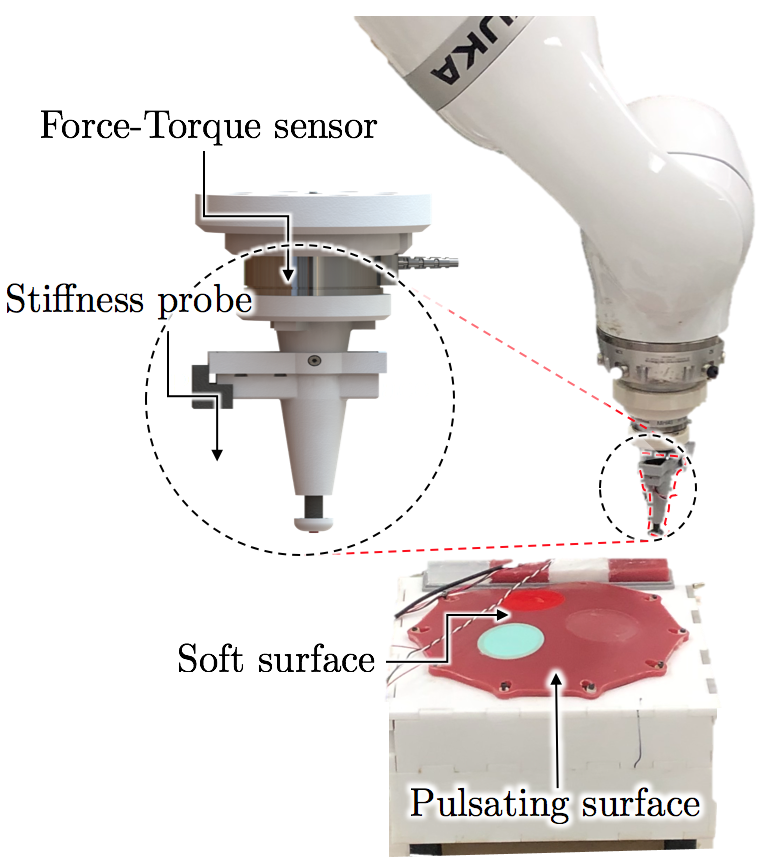}
\caption{Setup of the experimental setup. KUKA manipulator attached with a stiffness probe on a pulsating surface.}
\label{fig:setup}
\vspace{-20pt}
\end{figure}
 A comparative study on admittance and impedance control was done in \cite{Ott2010UnifiedControl}, where a unified approach was used to achieve better force tracking performance in a broader spectrum of the environmental stiffness. It was shown that admittance control is more suitable in compliant environments, while impedance is more suitable in stiff environments. But the stiffness should be known or observable to use the above-unified framework. In \cite{Rossi2016ImplicitMotion, Roveda2013Deformation-trackingEnvironments}, methods were presented for the estimation of the environment impedance that requires high-frequency input (position or force) to the end-effector through the manipulator. Though these methods are well-suited to industrial applications in stiff environments (e.g., grinding, polishing), they are not as feasible in applications such as the medical domain, or where there are spatial constraints or safety must be taken into account.  Moreover, in compliant environments, it cannot be assumed that the interaction environment is static or that stiffness is uniform along the indentation depth (in the direction of probing). \par
 Technical contributions of this paper can be listed as below:
 \begin{itemize}
 \item Design of a low-cost, low-inertia, safe stiffness identification probe for compliant environments which is independent of the robot configuration and motion.
 \item Fast convergence on material parameters and accurate identification of environmental stiffness using that probe
 \item Demonstrations of force control with motion compensation on a static and stationary compliant surface.
 \end{itemize}
\section{Related Work}
\vspace{-4pt}
Understanding the mechanical properties (e.g., stiffness)  of the environment is essential in force-controlled robotics applications such as robot teleoperation and autonomous robotic manipulation of soft materials for increased safety and performance.
Current medical practices use sophisticated techniques(CT, US, MRI) and other hand-held palpation devices \cite{Herzig2018APalpation, Faragasso2015Multi-axisPalpation}. However, these technologies cannot provide a direct measure of tissue elasticity, or they add a compliant element to the system where instabilities could amplify.   
\par
Model representation for contact force control has been explored widely, and simplistic models for sliding contact motion\cite{Howe1996PracticalManipulation} and, friction coefficients\cite{Fazeli2017ParameterContact} have been studied. These are developed for stiff contacts on a stationary environment where deformation of the contact body is comparable to or higher than the system (e.g., manipulator). For compliant contacts, impedance parameters vary spatially at a relatively high rate, and models have been developed to describe the behavior of such bodies\cite{Marhefka1999ASystems}. However, integration of non-linear models complicates control schemes as well as introduces numerical errors (e.g., from integration), which can degrade performance and introduce instabilities if compliance elements happen to be stiffer than the environment. \par
Many force control techniques have been developed in the framework of impedance control and direct force control\cite{H.AsadaRobotControl,Ott2010UnifiedControl,Hogan1985ImpedanceManipulation}. Robust control could be applied, as in \cite{Surdilovic2007RobustRobots}, or compliance parameters can
be estimated and compensated with an adaptive approach, as shown in \cite{Kroger2004AdaptiveExperiments,Mester1994AdaptiveRobots}. In these works, manipulator motion and estimated parameters are dependant on each other. As a result, the bandwidth of the excitation level is limited through the control input. 
In \cite{Rossi2016ImplicitMotion}, environment contact parameters are estimated on-line. It is integrated into the control scheme where the environment is stationary, rigid, and simplistic in geometry. Here, estimation and control are independent, where high-frequency sinusoidal input is added into the input, and frequency response is observed through a bandpass filter. It is well suited for a stationary environment where contact parameters do not vary quickly. In another approach, the stiffness parameters are estimated together with the normal and tangential directions of the contact surface and its friction coefficients \cite{Kikuuwe2005RobotImpedance}. It has an improved transition response of parameter estimation through the use of a forgetting factor. These works have shown that if the stiffness of the system is known and compensated for, higher force-tracking performance can be achieved along with increased stability. Hence, contact parameter estimation and compensation are critical in safety-critical applications. \par
Several approaches have been proposed for online parameter estimation: recursive least squares \cite{Love2002EnvironmentControl}, MRAC, indirect adaptive control \cite{Erickson2003ContactSystems}, EKF \cite{Roveda2013Deformation-trackingEnvironments,Verscheure2009IdentificationPayloads}, and algorithms based on an active observer \cite{Cortesao2007OnObservers}. In these works \cite{Kikuuwe2005RobotImpedance},\cite{Milek1995Time-VaryingIdentification}, RLS was used with linear and exponential time-varying forgetting factors to increase the transient response, and overall stability of it was discussed.  \par

% The overall system stiffness and relative environmental stiffness has an effect on the geometry of the environment. In almost all the other work, surface geometry is known a priori or simplifying assumptions have been made. In \cite{Kazerooni1989OnControl}, a generalized assumption on the geometry is made and analysed. 
% has been presented   and  to be  Where environment stiffness is considerably lower than the manipulator stiffness, the effect of geometrical uncertainty is not very critical. Even the robot compliance could be safely neglected. More to add...

\section{Proposed Work}
\subsection{Overall Outline}

The robot manipulator (KUKA IIWA LBR R820) is equipped with a force-torque sensor (ATI mini45) and with a  stiffness probing device ( as shown in Figure \ref{fig:stiffness_probe_figure}) for online stiffness estimation. Besides, a stability analysis is done to show the need for a stiffness adaptation while the environment is moving. A static and dynamic environment (Figure \ref{fig:setup}) is used for experimental validation. A dynamic environment is simulated using a pulsating silicone membrane that contains patches of silicone of different stiffness throughout the surface\footnote{For a clearer description, see the accompanying video}. \par

% The structure of this paper is as follows: first, the stiffness probe and the contact model used is discussed. The validation of the device with stiffness estimation and fast convergence of the parameters are then shown. Then, the control scheme of the overall is described and the stability and performance of it is assessed.  Finally, demonstrations of force control with motion compensation on a stationary and non-stationary compliant surface will be presented.

\subsection{Impedance Parameter Estimation Probe}
\begin{figure}[!t] 
\centering
\includegraphics[width=1.0\linewidth]{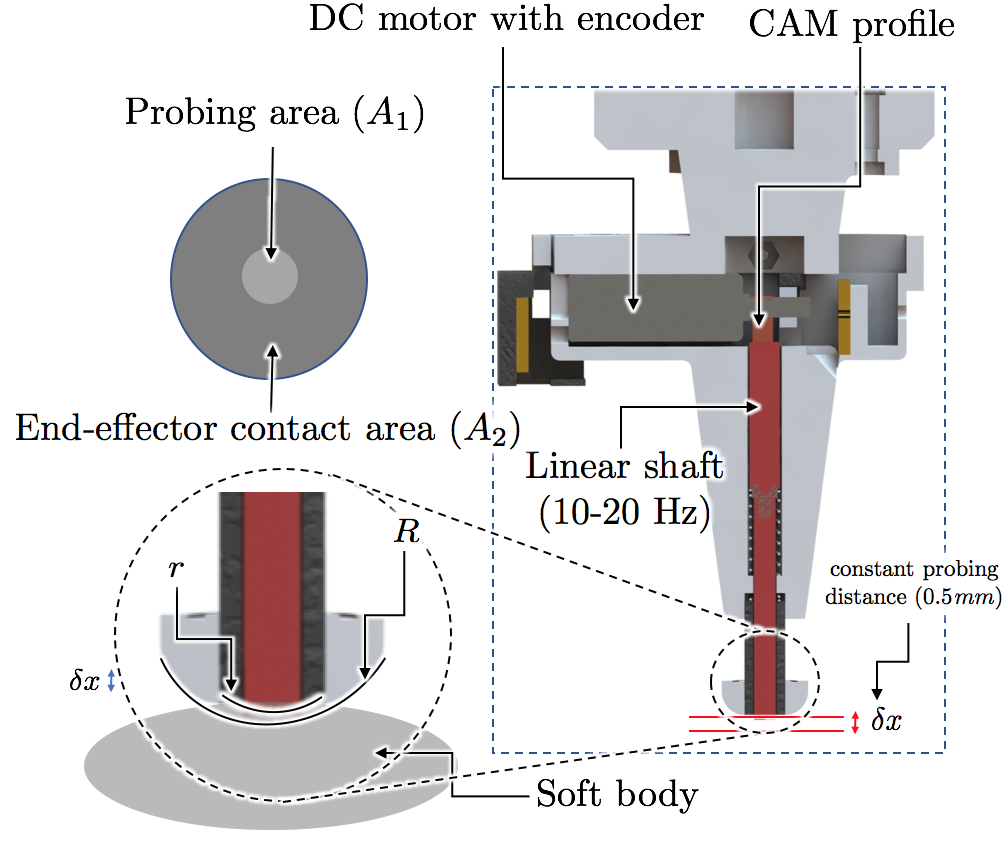} 
\caption{Graphical illustration of the proposed stiffness probe. $R$ and $r$ are the effective contact radii for the Eq.~(\ref{eq: hertz_model})}
\label{fig:stiffness_probe_figure}
\vspace{-10pt}
\end{figure}

% \begin{figure}[!h] 
% \centering
% \includegraphics[width=0.6\linewidth]{Figures/spring_model_1.png} 
% \caption{Stiffness Probe}
% \label{fig:stiffness_probe_figure}
% \vspace{-10pt}
% \end{figure}
A stiffness probing device attached to the end-effector of the manipulator (refer to Figure \ref{fig:stiffness_probe_figure}) is used to estimate the stiffness of the environment. Importantly, here, the parameter estimation is independent of the robot's current configuration and the underlying controls scheme.  This device is actuated to pulse at a rate of $20\text{Hz}$ with a probing depth of $0.05\text{mm}$ (manually adjustable). Since the excitation input is directly the probing displacement, this is well suited for low-impedance environments.  Moreover, the convergence of time-varying parameters is faster with high-frequency persistent excitation \cite{Milek1995Time-VaryingIdentification}. The force response of the excitation is captured through the FIR band-pass filter ($8-18\text{Hz}$) from the force sensor. The position of the linear shaft and the corresponding force sensor reading is recorded and used for the online stiffness estimation with a recursive least squares (RLS) with an adaptive forgetting factor.
\subsection{Overall Model Representation}
\begin{figure*}[t!] 
\centering
\vspace{2pt}
\includegraphics[width=.9\linewidth]{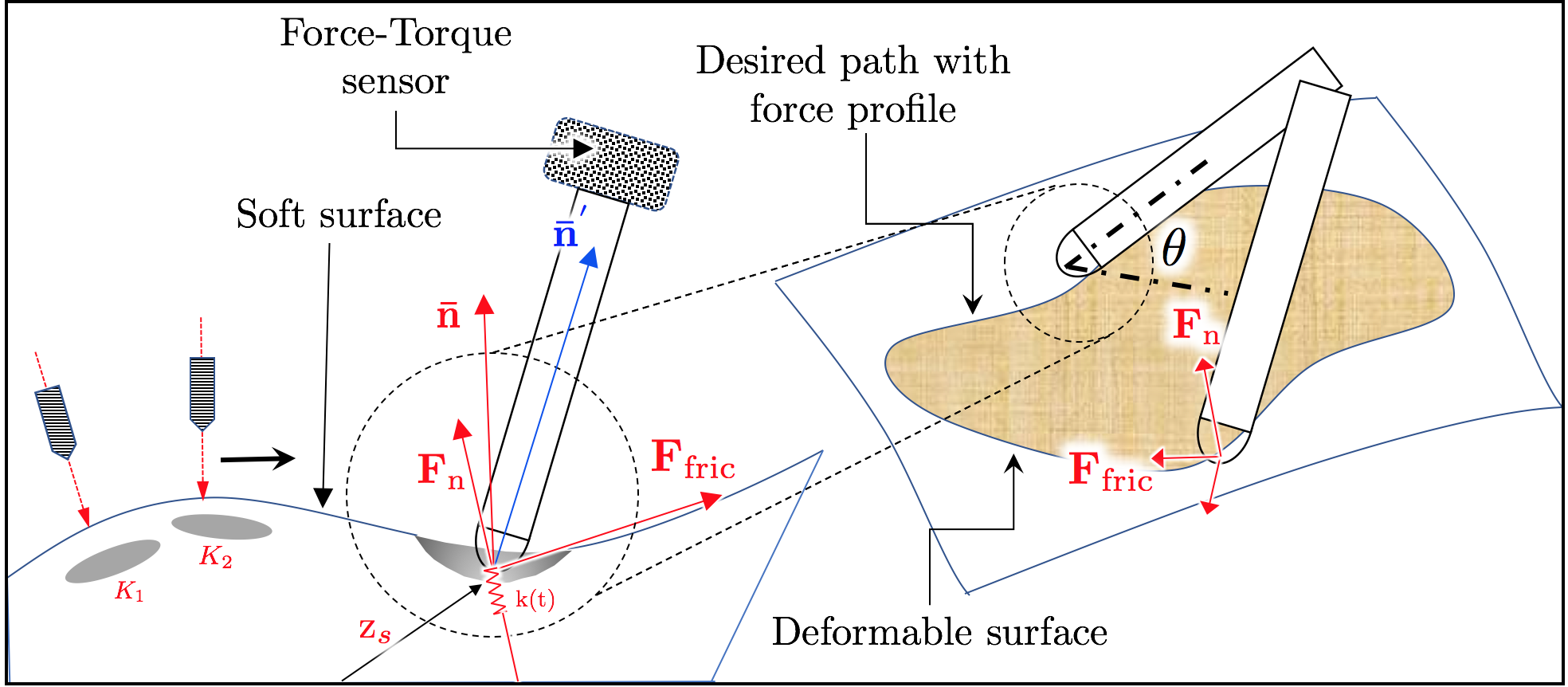} 
\caption{Illustration of the contact model. Tool in contact with the soft surface with varying contact properties while moving. Mathematics symbols used here are refereed in the text. Note the linear spring in the direction of $\mathbf{\bar{n}}$}
\label{fig:contactModelFig}
\vspace{-20pt}
\end{figure*}

Contact models for compliant bodies differ from stiff environments where stiffness is uniform throughout the body, and uncertainty is low. For a silicone gel model, indentation depth and force relation could be represented as in Eq.~(\ref{eq: hertz_model}) \cite{Johnson1985ContactMechanics}. It shows the non-linear contact behavior for a deformable (soft) body. The local stiffness estimated for the force controller is \textbf{A} from Eq.~(\ref{eq: hertz_model}), where $K_E$ cannot be assumed to be uniform throughout the indentation depth. If the underneath layer (e.g. stiff nodule) is slightly different in stiffness ($K_2$ as depicted in Figure \ref{fig:contactModelFig}), overall, lumped stiffness varies significantly. \par
As the effective local stiffness varies with the contact area, the stiffness estimated by the probe has to be scaled by a factor which is $c = \big(R/r\big)\bar{K}_E$  (as in Figure \ref{fig:stiffness_probe_figure} and the definition of Young Modulus). \remark{As $A_1 \rightarrow{A_2}$, $K_E \rightarrow{\bar{K}_E}$, where $\bar{K}_E$ is the estimated stiffness through the probe. $A_1$ and $A_2$ are effective contact areas of probe and tool.} 
\begin{equation}
    F = \frac{4\sqrt{\bar{R}}\hspace{2pt}E\delta^{1.5}}{3\hspace{2pt}(1-\nu^{2})} \quad\text{and}\quad \frac{\partial{F}}{\partial{\delta}} = \underbrace{\Bigg(\frac{2\sqrt{\bar{R}}\hspace{2pt}E}{\hspace{2pt}(1-\nu^{2})\sqrt{\delta}}\Bigg)}_{\textbf{A}}\delta
    \label{eq: hertz_model}
\end{equation} 
where, $E$ is the young modulus, $\delta$ is the indentation depth, $\bar{R}$ is the the effective indentor radius, $\nu$ is the poisson's ratio \footnote{For silicone material $\nu \sim 0.5$}.
\par  
Figure \ref{fig:contactModelFig} shows the generalized contact model that is used in this paper. $\mathbf{\bar{n}}$ represents the average surface normal (net force vector as a result of the overall local deformation, as shown in Figure \ref{fig:contactModelFig}) while $\mathbf{\bar{n}^{'}}$ represents the current orientation of the end-effector. The environment is simplistically represented as a one dimensional linear spring ($\text{k(t)}$) in the vector direction $\mathbf{\bar{n}}$. It could be extended to higher dimensions if desired. The force admittance controlled direction of the end-effector movement is the projection of the end-effector movement to the average surface normal vector ($\mathbf{\bar{n}}$). Sliding friction can be assumed to be acting in the moving direction. Then, the generalized force model can be written as:
\begin{multline}
    \mathbf{F}(t) = - K_E(\mathbf{x_s},\mathbf{y_s},t)\hspace{2pt}\mathbf{\bar{n}(t)^{}\bar{n}(t)^{T}}\hspace{2pt}[\mathbf{x_{e}}(t) - \mathbf{z_{s}}(t)] \\+ K_f\proj_{\mathbf{\bar{n}}^{\perp}}(\dot{\mathbf{x}}_e(t) - \dot{\mathbf{z}}_s(t)) 
\end{multline}
% \begin{equation*}
%     \text{Linear Spring Vector} \rightarrow{\mathbf{\bar{n}}} 
% \end{equation*}
 $\mathbf{\bar{n}}$ can be approximated by the force vector perceived at the end-effector by the force sensor. To mitigate the noise in the force reading, a FIR low pass filter (with $8\text{Hz}$ cutoff) is applied. \assumption{We assume the surface geometry and stiffness variation through out the surface are sufficiently continuous. \label{assum: cont}} \\
Let $\mathbf{N} = \mathbf{\bar{n}}(t)\mathbf{\bar{n}}^T(t)$ and $\mathbf{N_f} = \proj_{\mathbf{\bar{n}}^{\perp}}$
Then, 
\begin{multline}
    \mathbf{F}(t) = - K_E(\mathbf{x_s},\mathbf{y_s},t)\hspace{2pt}\mathbf{N}\hspace{2pt}[\mathbf{x_{e}}(t) - \mathbf{z_{s}}(t)] +\\ K_f\hspace{2pt}\mathbf{N_f}\hspace{2pt}(\dot{\mathbf{x}}_e(t) - \dot{\mathbf{z}}_s(t))
    \label{eq:force_1}
\end{multline} 
For simplicity, in this study, friction is omitted since deformable soft bodies tend to have lower friction coefficients in comparison to the other mechanical properties, and we do not focus on modulating frictional forces. In cases where contact surfaces are not sufficiently smooth, this would induce significant errors on the average surface normal vector while in motion. 
In \cite{Rossi2016ImplicitMotion}, a similar analysis on variations due to stiffness and geometry was presented. Similarly, by taking the total derivative of Eq.~ (\ref{eq:force_1}), 
\begin{equation}
    \frac{d\mathbf{F}(t)}{dt} = -\nabla_{xyz}\mathbf{F}(t)\boldsymbol{\upsilon}  - \frac{\partial\mathbf{F}(t)}{\partial t}
    \label{eq:force_1}
\end{equation} 
\vspace{2pt}where, $\boldsymbol{\nabla}* = \big[\sfrac{\partial}{\partial{\mathbf{x}}}\hspace{3pt} \sfrac{\partial}{\partial{\mathbf{y}}}\hspace{3pt} \sfrac{\partial}{\partial{\mathbf{z}}}\hspace{3pt}
\sfrac{\partial}{\partial{\mathbf{x_s}}}\hspace{3pt}
\sfrac{\partial}{\partial{\mathbf{y_s}}}\hspace{3pt}
\sfrac{\partial}{\partial{\mathbf{z_s}}}\big]$,\\ \vspace{1pt}and 
$\boldsymbol{\upsilon}(x,y,z,x_s,y_s,z_s) = \big[\dot{x}\hspace{2pt} \dot{y}\hspace{2pt} \dot{z}\hspace{2pt}\dot{x}_s\hspace{2pt} \dot{y}_s\hspace{2pt} \dot{z}_s\big]$ 
\vspace{5pt}
\\where, $K_E$, is the environment stiffness, $\mathbf{z_{s}} \in \R^{3\times 1}$ is the surface cartesian position with respect to the world frame, and $\mathbf{x_{e}} \in \R^{3\times 1}$ is the cartesian position of the robot end-effector. By further expanding,
\begin{multline}
   \frac{d\mathbf{F}(t)}{dt} = \Big[\underbrace{(\boldsymbol{\nabla}K_E)}_{R_1}\hspace{2pt}\mathbf{N}\hspace{2pt}[\mathbf{x_{e}}(t) - \mathbf{z_{s}}(t)]  \hspace{4pt}+ \\K_E\underbrace{(\boldsymbol{\nabla}\mathbf{N})}_{R_2}[\mathbf{x_{e}}(t) - \mathbf{z_{s}}(t)]  \hspace{4pt}+\\ K_E\hspace{2pt}\mathbf{N}\hspace{2pt}\underbrace{(\boldsymbol{\nabla}[\mathbf{x_{e}}(t) - \mathbf{z_{s}}(t)])}_{R_3}\Big]\boldsymbol{\upsilon}\hspace{2pt}-\\ \Big[ (K_E(t)\mathbf{N})\Big] (\dot{\mathbf{x}}_e - \dot{\mathbf{z}}_s)
   \label{eq:model_variations}
\end{multline}

Eq.~(\ref{eq:model_variations}) gives the overall combined spatial and time variation along $\mathbf{\bar{n}}$ vector. With the Assumption (\ref{assum: cont}), $R_2$ term (variation of the surface normal) in the Eq.~(\ref{eq:model_variations}) could be regarded as a quasi-static variable. Then Eq.~(\ref{eq:model_variations}) could be reduced to Eq.  ~(\ref{eq:model_delta}),

By further grouping the terms together, and
\\if ($\mathbf{x_{e}}(t) - \mathbf{z_{s}}(t)) \rightarrow{\boldsymbol{\delta}}(t)$, Eq.~(\ref{eq:model_variations}) could be written as: 
\begin{multline}
   \frac{d\mathbf{F}(t)}{dt} = -\Big[(\boldsymbol{\nabla}K_E)\hspace{2pt}\mathbf{N}\hspace{2pt}\boldsymbol{\delta}(t)  \hspace{4pt}+\\\hspace{4pt} K_E\hspace{2pt}\mathbf{N}\hspace{2pt}(\boldsymbol{\nabla}\boldsymbol{\delta}(t))\Big]\boldsymbol{\upsilon}- ( K_E(t)\mathbf{N})\dot{\boldsymbol{\delta}}(t)
   \label{eq:model_delta}
\end{multline}
By having $\mathbf{N}\hspace{2pt}\boldsymbol{\delta}(t) = \delta(t)$ (along the surface normal vector),
and taking the laplace transfomation of Eq.~(\ref{eq:model_delta}),
\begin{multline}
    s\mathcal{L}(\mathbf{F}(s)) = -\Big[(\boldsymbol{\nabla}K_E)\hspace{2pt}{\mathcal{L}(\delta}(s))  \hspace{4pt}+\hspace{4pt} K_E\hspace{2pt}\hspace{2pt}(\boldsymbol{\nabla}\mathcal{L}(\delta(s))\Big]\hspace{4pt}-\\s\hspace{2pt}K_E(t))\mathcal{L}(\delta(s))
\end{multline}
If input is $\delta(t)$ and output is $\mathbf{F}(t)$, then, transfer function could be written as:
\begin{multline}
    \frac{\mathcal{L}(\mathbf{F}(s))}{\mathcal{L}(\delta(s))} = -\Big[\frac{1}{s}(\boldsymbol{\nabla}K_E)\hspace{2pt}\mathbf{N}\hspace{2pt}  \hspace{4pt}+\hspace{4pt} \frac{K_E\hspace{2pt}\mathbf{N}\hspace{2pt}(\boldsymbol{\nabla}\mathcal{L}(\delta(s)}{s\mathcal{L}(\delta(s))})\Big]\hspace{4pt}-\\\hspace{2pt}K_E(t)\mathbf{N})
    \label{eq:transfer_function}
\end{multline}

At sufficiently high frequencies, as the first two terms  approach zero, the Eq.~(\ref{eq:transfer_function}) could be approximated as:
\begin{equation}
    \frac{\mathcal{L}(\mathbf{F}(s))}{\mathcal{L}(\delta(s))}  \approx K_E(t)\mathbf{N}
    \label{eq:transfer_function_reduced}
\end{equation}

In \cite{Love2002EnvironmentControl,Rossi2016ImplicitMotion}, a similar approach was followed to estimate the surface stiffness. Excitation inputs ($\sim3\text{Hz}$)  were augmented with a high frequency sinusoidal positional or force input to the reference input. It is limited by the manipulator bandwidth and could be hazardous as the manipulator impedance, and natural resonance frequency could vary with the configuration. For a dynamic environment that is non-stationary, this approach cannot be used unless an excitation frequency higher than the highest frequency component of the environmental disturbance is used. \par
The stiffness probe presented in section B is based on a constant positional excitation input at a sufficiently high frequency ($\sim20\text{Hz}$). For instance, in medical domain applications where stiffness varies spatially and with time, such as chest/heartbeat force compensation. 
\section{Methods}
\subsection{Performance and Stability Analysis}
The performance of force control can be assessed in terms of performance and stability. For safety-critical applications, performance metrics such as maximum overshoot and oscillations are essential in addition to the instabilities that could occur in the system. In indirect force control, the end-effector movement is computed through a controller with a proportional gain that represents the environment stiffness. For instance, if the same controller gains used for a very stiff environment are used for a soft environment, performance would be inadequate. On the other hand, if the controller gains used for a soft environment are used for a stiff environment, high amplitude oscillations or instabilities could occur, which are undesirable. More comprehensive stability analysis of this nature is done in \cite{Kazerooni1989OnControl}. \par
\begin{figure*}[t!]
\centering
  \includegraphics[width=.95\linewidth]{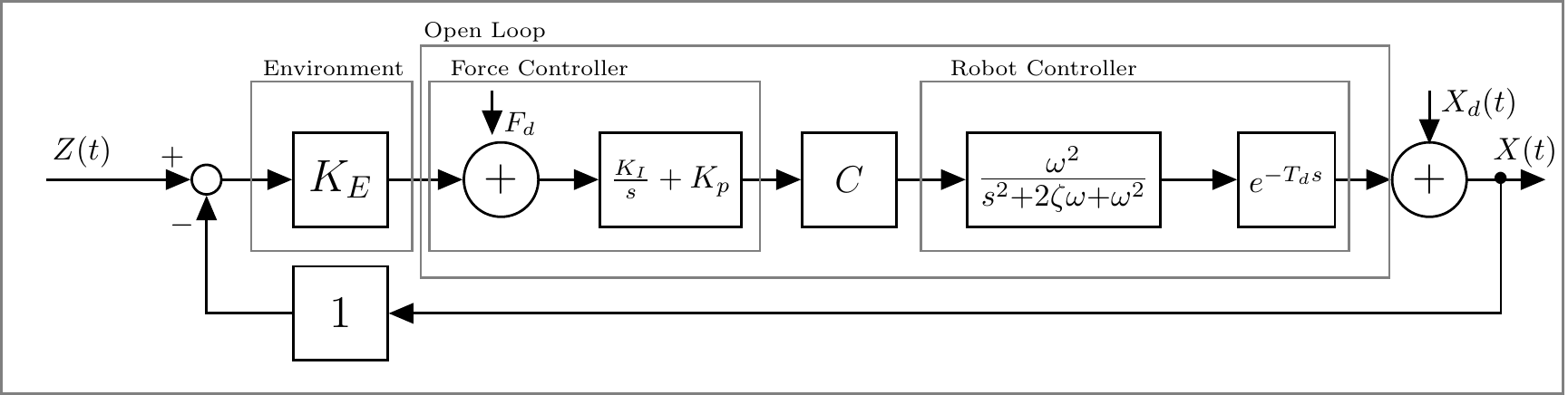}%<-- Replace 'Preview' with your figures file's name
  \caption{Force Control Flow Diagram(closed loop). $Z(t)$ is the environmental position, $F_d$ is the desired force, and $X_d(t)$ is the desired robot reference position. $\omega = 300$, $\zeta = 0.9$. This represents a 1-DOF system in Cartesian space. Corresponding joint commands are generated in section \ref{sec:joint control}}
  \label{simple_control_model}
  \vspace{-15pt}
\end{figure*}
Figure \ref{simple_control_model} shows the environmental-manipulator interaction model for one dimensional case in the Cartesian space. \footnote{It can be extended to higher dimensions as desired with the assumption that degrees are uncoupled}. Here, $Z(t)$ is the displacement of the environment and $X(t)$ is the resulting Cartesian position along the average surface normal($\mathbf{\bar{n}}$). $F_d$ is the desired force value to be tracked, and $C$ is the estimated compliance of the environment ($C=K_E^{-1}$). A Proportional-Integral (PI) controller is used as the force controller\footnote{Gains were tuned to have a critically damped system in the ideal case where estimated stiffness is exactly the real}, and system delays due to joint friction and the low-pass FIR filter are lumped into one term, ($T_d = 0.1\text{s}$). The delay is approximated using a $2^{nd}$ order $pad\acute{e}$ approximation. \par
\begin{figure}[b!] 
\centering
\vspace{1pt}
\includegraphics[width=.95\linewidth]{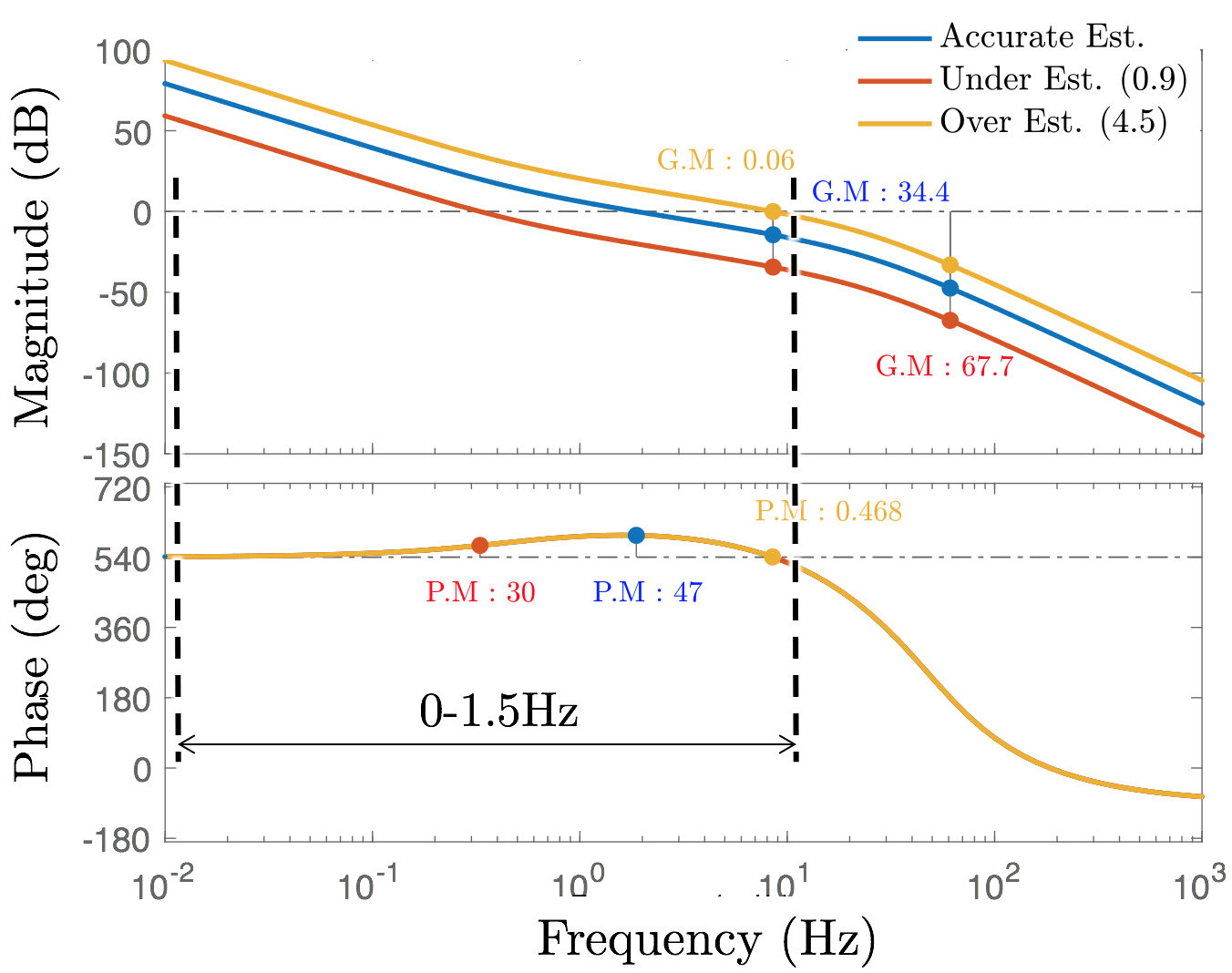}
\caption{Bode plot of the open loop plant. G.M : Gain Margin, P.M : Phase Margin.} 
\label{fig:bode}
% \vspace{-20pt}
\end{figure}
Figure \ref{fig:bode} shows the open-loop bode of the system with the delay. Frequency responses for under, over, and ideal stiffness estimations are plotted together with gain and phase margins. It is observable that the gain shifts to be positive in magnitude when the phase is $180\deg$ where stiffness is underestimated, which suggests the closed-loop system becomes unstable. Over-estimation of the impedance parameters can be compensated for without instabilities, while underestimation could drive the system to instability due to the delay in the system. For dynamic performance, in addition to stability analysis, response time and oscillatory behavior must be taken into consideration. In the overestimated case, response time tends to be slow, which deteriorates dynamic performance.  
% This affects the bandwidth it could handle 
% \begin{figure*}[t!]
% \centering
%   \includegraphics[width=.80\linewidth]{Figures/flow_3.pdf}%<-- Replace 'Preview' with your figures file's name
%   \caption{Force Control Flow Diagram}
%   \label{overall_flow}
%   \vspace{-20pt}
% \end{figure*}
% \begin{figure}[h!] 
% \centering
% \includegraphics[width=.75\linewidth]{Figures/rl.png}
% \caption{Root locus plot. Marked x's are poles in the ideal case. The blue line represents poles when the system goes unstable (at least one pole goes unstable). The unstable zone is shaded in grey.}
% \label{fig:root_locus}
% \vspace{-20pt}
% \end{figure}

\subsection{Contact Parameter Identification through a Probe}

For compliant bodies, the contact parameter model is represented through non-linear contact models (e.g. Hertz model\cite{Gilardi2002LiteratureModelling,Pappalardo2016Hunt-CrossleySurgery}). For control simplicity, the contact environment impedance is modelled as a simple spring-damper system that could be represented as Eq.~(\ref{eq:spring-damper})
\begin{equation}
    \mathbf{F}_n = K(x,y,t)\big(\mathbf{x_e - x_s}\big) + D(x,y,t)\big(\mathbf{\dot{x}_e - \dot{x}_s}\big) \label{eq:spring-damper}
\end{equation}

where, $\mathbf{\dot{x}}_s$ is the environment's surface's moving velocity and $\mathbf{x}_e$ is the robot's cartesian velocity projected in the direction of the surface average normal.

A bilinear transformation is then used to transform Eq.~(\ref{eq:spring-damper}) into its discrete-time counterpart, which results in,
\begin{equation}
    F_n = \Bigg(D_E\Bigg(\frac{2}{T}\Bigg)\Bigg[\frac{1-z^{-1}}{1+z^{-1}}\Bigg] + K_E\Bigg)\delta
\end{equation}
By recognizing that $z^{-1}$ represents
a backward shift of one step in the time domain, and letting k describe the time-step index, the equivalent discrete time equation is:
\begin{equation}
\small
    F_k + F_{k-1} = \underbrace{\frac{1}{T} \Big(2D_E+TK_E\Big)}_A\delta_k + \underbrace{\frac{1}{T}\Big(-2D_E+TK_E\Big)}_B\delta_{k-1}
\end{equation}
By putting this into a regressor form,
\begin{equation}
    y_k = \boldsymbol{\phi}_{k} \boldsymbol{\theta}^{T}_k
\end{equation}
with,
\begin{equation*}
    \mathbf{\phi}_k = \begin{bmatrix}
    \delta_k &
    \delta_{k-1} \\
    \end{bmatrix}
    \quad\text{and}\quad 
    \mathbf{\theta}_k = \begin{bmatrix}
    A & B \\
    \end{bmatrix}
\end{equation*}

\begin{equation}
    \bar{\mathbf{P}}_{k+1} = \big(\mathbf{P}_k^{-1} +  \boldsymbol{\phi}^T\boldsymbol{\phi}\big)^{-1} : \mathbf{P}_0 > 0
\end{equation}
\begin{equation}
    \bar{\boldsymbol{\theta}}_{k+1} = \bar{\boldsymbol{\theta}}_{k} + \bar{\mathbf{P}}_{k+1}\boldsymbol{\phi}_k^T\big(\mathbf{y}_k-\boldsymbol{\phi}_{k} \boldsymbol{\bar{\theta}}^{T}_k\big)
\end{equation}
\begin{equation}
    \mathbf{P}_{k+1} = \mathit{f}(\bar{\mathbf{P}}_{k+1})
    \label{pk}
\end{equation}
$\bar{\boldsymbol{\theta}}_{k}$ is the 
where, $\mathcal{F}(\bar{\mathbf{P}}_{k+1}) = \mu_k\mathbf{P}_{k+1} + g_k\mathbf{J}_k$\\
where $\mu$ ($\mu_k > 0,\mu_k < 1  \hspace{2pt} \forall \hspace{2pt}k$) is the forgetting factor and $J_k$ is a positive definite matrix. ($g_k < \bar{g}, g_k > 0$). Analysis on stability for time varying $g_k$ and $\mu_k$ is done in \cite{Milek1995Time-VaryingIdentification}. It is shown that the stability of the estimator can be guaranteed if the estimator parameters ($g_k$ and $\mu_k$) are bounded and within the constraints. 

\subsection{Adaptation through Forgetting}
To have a responsive on-line estimation of the current stiffness, co-variance of the estimation has to adapt to the current observation error. We used an adapting forgetting factor ($\mu_k$), which is a function of the current measured observation error. By this method, when the error is small, the estimation would be less noisy and takes the history into account while it forgets faster when the error is larger to adapt to the current perceived stiffness. The adaptation law $\mu_k$ used is:
\begin{equation}
    \mu_k = \tanh{\big(|\mathbf{y}_k-\boldsymbol{\phi}_{k} \boldsymbol{\bar{\theta}}^{T}_k|\big)}
\end{equation}
\subsection{KUKA Manipulator Joint Control}
\label{sec:joint control}
 Joint references are generated as an input and corresponding torque output values are computed through the internal controller of the robot. KUKA FRI \cite{KukaManual} allows to command joint references at rate of $1000\text{Hz}$. (See Figure \ref{simple_control_model}). Instantaneous joint movement is calculated as:
\begin{equation}
    \delta \mathbf{q} = \mathbf{J}^{\dagger}_m(\mathbf{q})\hspace{3pt}C\hspace{3pt}(\delta{\mathbf{F}})\hspace{3pt}\footnote{$\mathbf{J_m^{\dagger}}$ represents the Moore–Penrose inverse}
\end{equation}
where, $\mathbf{J}_m \in \boldsymbol{\mathbb{R}}^{7 \times 7}$, is the analytical Jacobian of the manipulator at the current configuration $\mathbf{q}$. $\delta{\mathbf{F}}$ is the current error residual from the PI force controller (shown in Figure \ref{simple_control_model})
, and $\delta \mathbf{q}$ is integrated over time and given to the robot as the reference joint positions.

% \begin{figure}[!h] 
% \centering
% \includegraphics[width=0.5\linewidth]{Figures/depth.png} 
% \caption{Stiffness variation along the depth of indentation}
% \label{fig:depth}
% \vspace{-20pt}
% \end{figure}
% \subsection{Estimation of the surface movement}
% The surface movement of the environment is estimated using a extended kalman filter. The state is defined as: 
% \begin{equation}
%     X = [F, z, \dot{z}, x, \dot{x}] 
% \end{equation}
% \begin{equation}
%     \textit{f}(\mathbf{X}) = 
%     \begin{bmatrix} 
%         K_E(t)(z-x) + D_E(t)(\dot{z}-\dot{x})+\epsilon_x\\
%         \dot{z} \\
%         \epsilon_z \\
%         \dot{x} \\
%         \epsilon_x
%     \end{bmatrix}
% \end{equation}

\section{Experiments and Results}
\begin{figure}[b!] 
\centering
\includegraphics[width=0.9\linewidth]{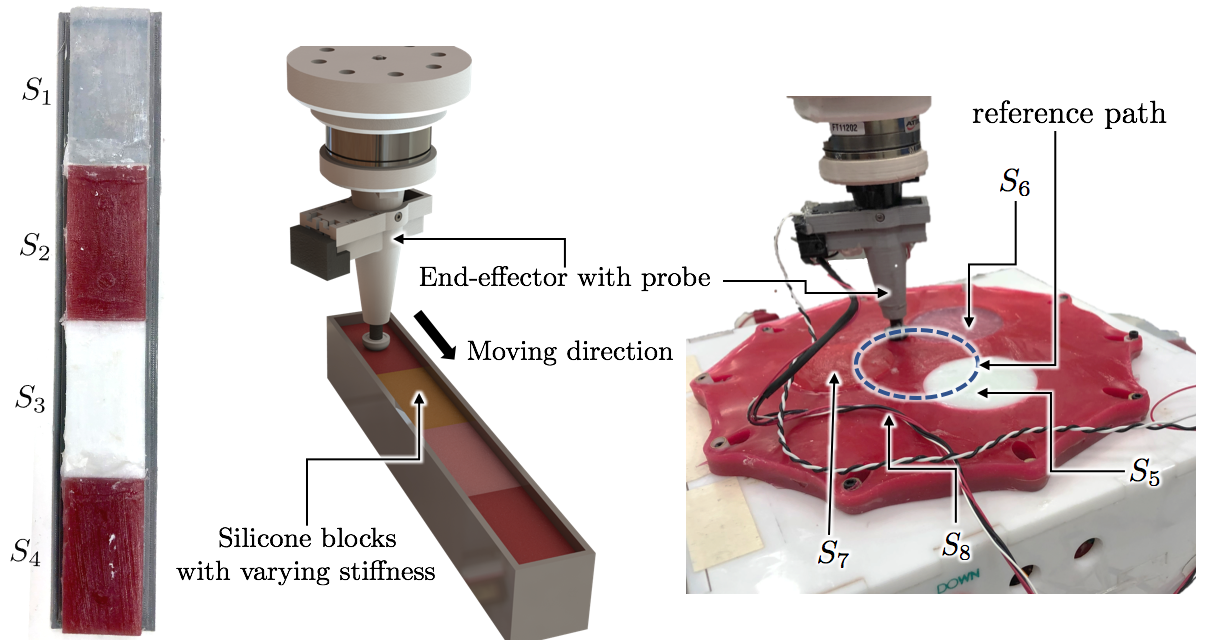}
\caption{Experimental Setup. (left) setup for the static environment. (right) setup for the non-stationary environment which is pulsating at 0.5Hz}
\label{fig:stiffnessSetup}
\vspace{-10pt}
\end{figure}
\subsection{Validation of Environment Contact Parameter Estimation}
\label{sec:stiffness_validation}
\begin{figure}[b!] 
\centering
\vspace{5pt}
\includegraphics[width=0.8\linewidth]{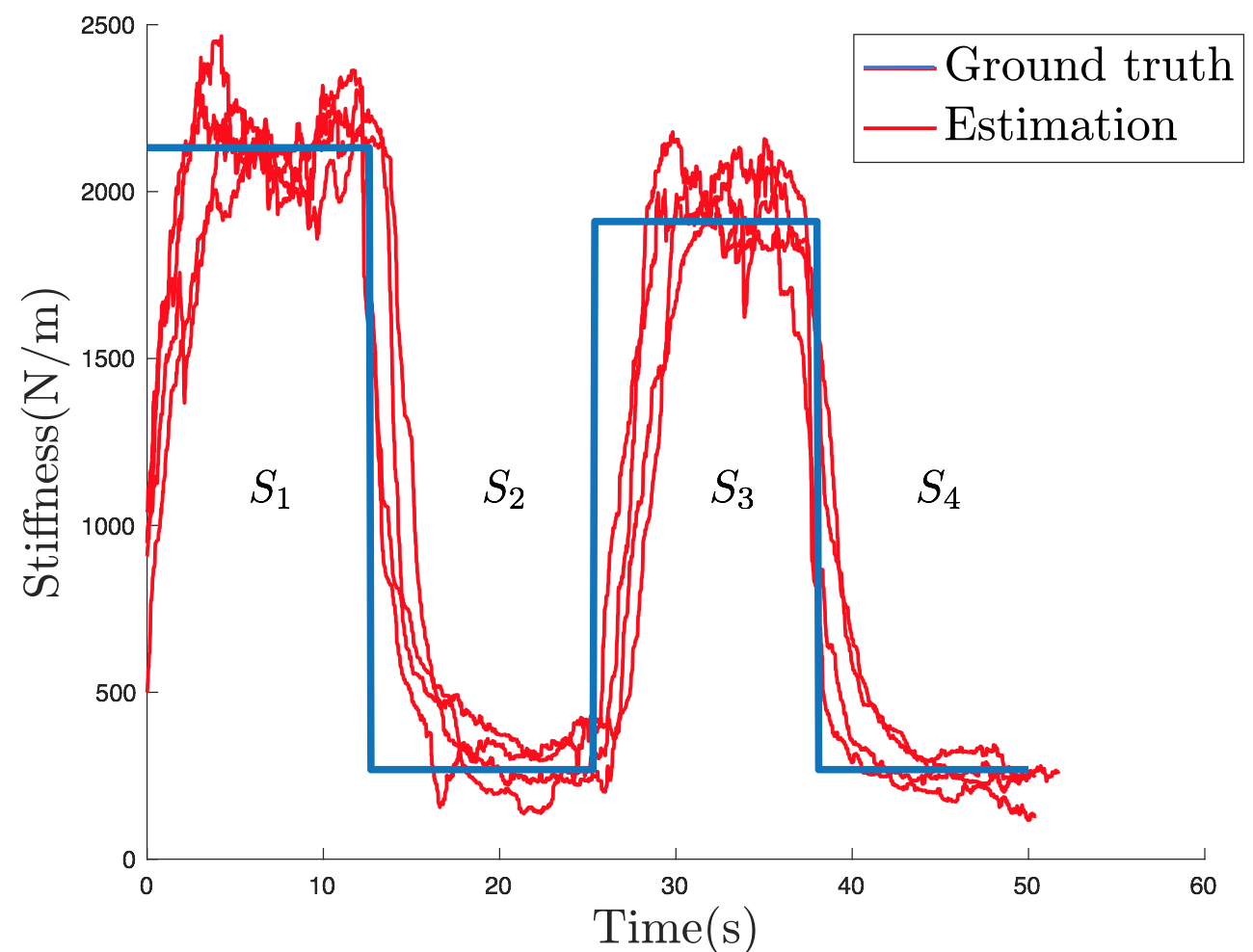} 
\caption{Stiffness estimation validation. Solid blue line is the ground truth. Solid red line is the estimated value}
\label{fig:StiffnessEstFig}
\vspace{-15pt}
\end{figure}

A validation test was performed with four silicone cubes with varying stiffness attached in series to validate the accuracy and convergence properties of the stiffness probe, shown in Figure \ref{fig:static_slider}. The end-effector of the manipulator was moved along a straight line while estimating the stiffness. The resulting plot for five consecutive trials is shown in Figure \ref{fig:StiffnessEstFig}.  The ground truth of the stiffness was measured using cyclic testing on the INSTRON$^ \copyright$ with 0.1mm probing displacement. Impedance parameters were computed from least squares (LS) estimation on the data collected for a specific time. As compliant surfaces (e.g., silicone gel) tend to be non-uniform, a stochastic behavior in the estimation could be observed as well. It could be minimized by having a smaller $\mathbf{J}_k$ in equation \ref{pk} at the cost of a slower response rate. 

\subsection{Force Control on a Static Surface}

\begin{figure}[b!] 
\centering
\vspace{4pt}
\includegraphics[width=0.9\linewidth]{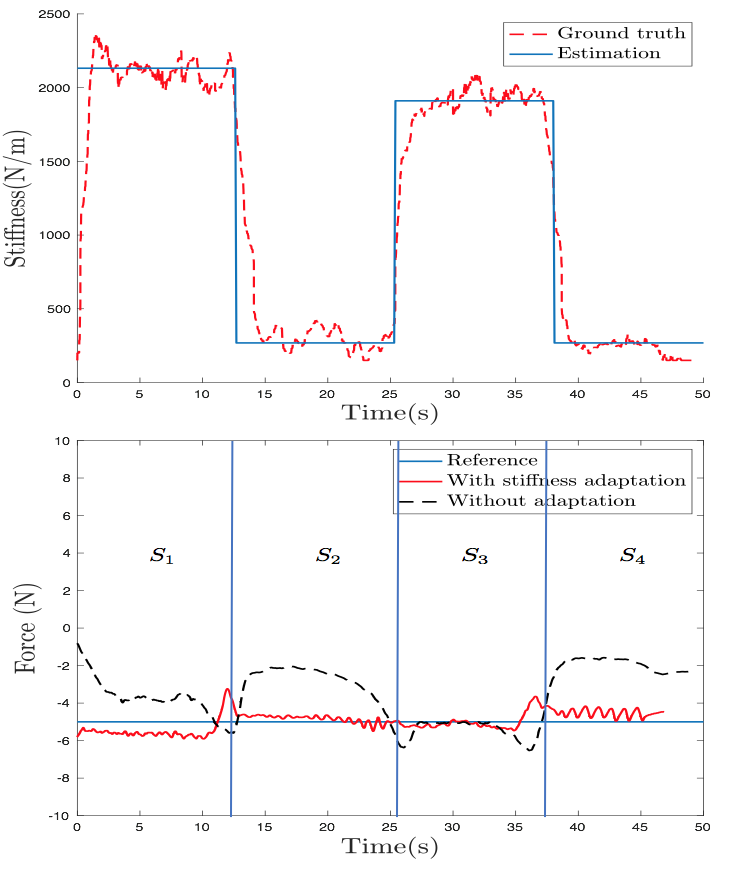} 
\caption{Sliding on a static surface with active stiffness estimation and force modulation}
\label{fig:static_slider}
% \vspace{-20pt}
\end{figure}
% \Reviewer{From Fig. 9 the author shows the result of the
% non-stationary environment, however, due to the pulsating
% surface, the stiffness of the deformed silicon should be
% measured and shown with the result as Fig 8.}

Force modulation on a static environment and validation of the stiffness probe was done on the same setup (section \ref{sec:stiffness_validation}). In this case, the force controller was active, and the reference force was set to 5N on the surface. Results are shown in Figure \ref{fig:static_slider}. In the case where stiffness adaptation was not used, $K_I$ and $K_P$ gains of the force controller were tuned such that the system is not unstable in the stiffest part of the silicone slider (as shown in Figure \ref{fig:static_slider}). This guarantees system stability.\par
It is observable that while stiffness adaptation is active, the controller is able to track the reference, as opposed to the alternative case. It should also be noted that, since transition points between areas of differing stiffnesses are not perfectly continuous (Figure \ref{fig:stiffnessSetup}), some overshoot is seen at those transitions.
\subsection{Force Control on a Non-stationary surface} 
\begin{figure}[t!] 
\centering
\vspace{2pt}
\includegraphics[width=0.85\linewidth]{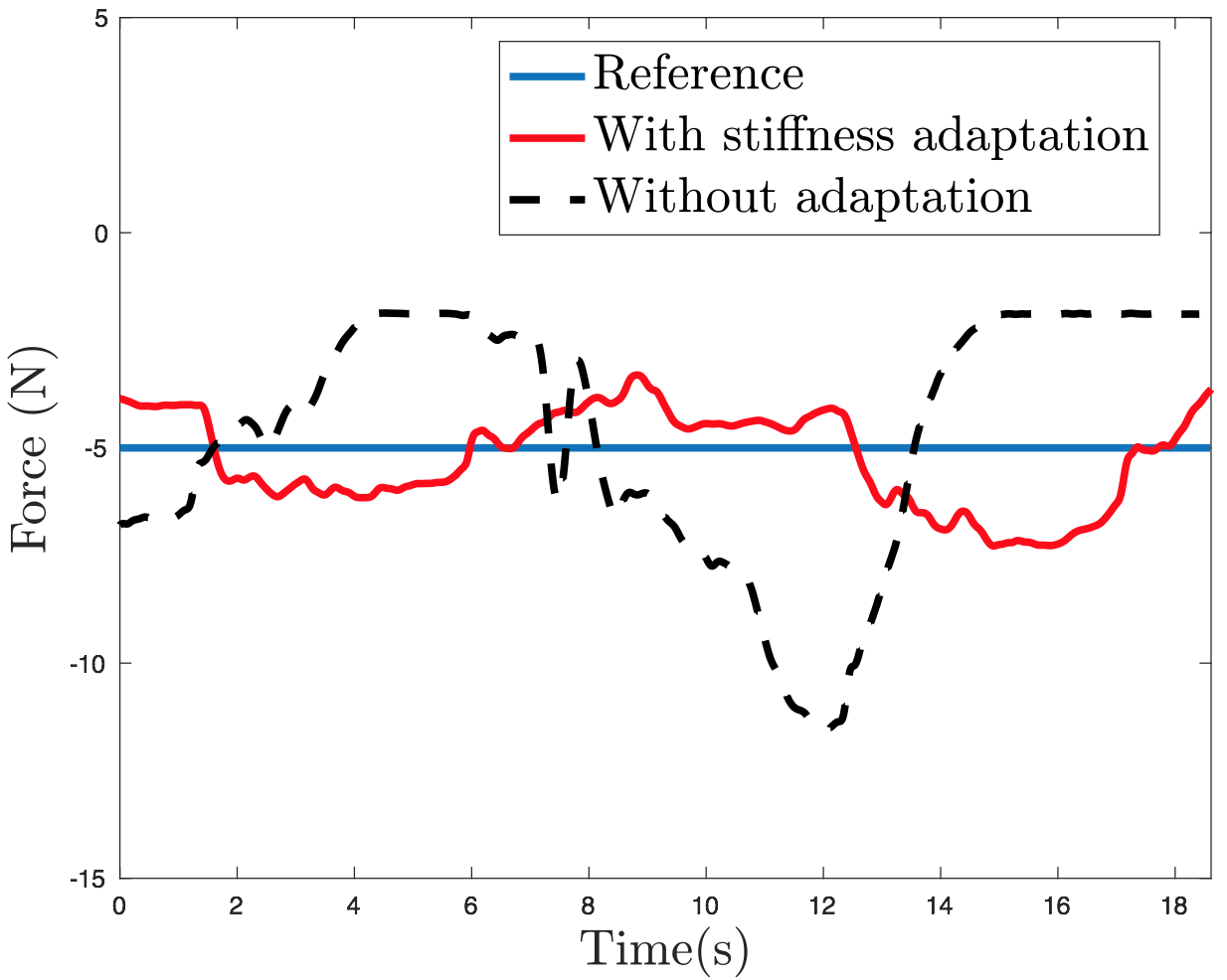} 
\caption{Force modulation on a non-stationary environment with pulsation at 0.5Hz}
\label{fig:dyn_force_control}
\vspace{-15pt}
\end{figure}

Simultaneous stiffness identification and force modulation on a non-stationary surface were performed to show the performance of the stiffness estimation and performance increase in force modulation. Setup for the experiment is shown in Figure \ref{fig:stiffnessSetup} (right). The pulsating surface is made to pulsate at a frequency of $0.5\text{Hz}$. Surface contained areas of different stiffness as labeled by $\text{S}_5$, $\text{S}_6$, $\text{S}_7$, $\text{S}_8$ . Ground truth of the stiffness value was not calculated in the dynamic case due to practical difficulties in using the INSTRON$^\copyright$ with the pulsating motion. Since the Compliant surface is made to pulsate\footnote{See the accompanying video for clarification}, and simultaneous stiffness estimation and force modulation were performed. Figure \ref{fig:dyn_force_control} shows the dynamic force modulation comparison when there is adaptation and no adaptation. It is to be noted that, when the environment is not stationary, the force controller alone cannot compensate for the surface movement. \cite{Dominici2014ModelSurgery} suggested an active observer to generate a feed-forward term to compensate for the environmental disturbance. A similar strategy was not followed in this paper since the focus was on the performance improvement by stiffness estimation. From figure \ref{fig:dyn_force_control}, it is observable that with stiffness estimation, force tracking performance is improved but not as well as in a static environment.
% \subsection{Estimation of the surface movement}

% \subsection{Comparison of results}

%Need more explanation here 
% \section{DISCUSSION}
% Results doesn't show perfect reference tracking and estimation(figure \ref{fig:StiffnessEstFig}). This is due to the fact that soft bodies are not uniform on the surface finish. 
\section{Conclusion and Future Work}
 This paper demonstrates online environmental stiffness identification and force modulation on both compliant stationary and non-stationary environments. Mechanical properties of compliant surfaces tend to be stochastic in nature and vary significantly with time and space. The proposed method for stiffness identification, which is independent of the manipulator configuration, has shown successful estimation and enhanced the performance of admittance force control. 
 \par 
 Future work of this will look into the estimation of environment motion and using a high-level trajectory optimizer that will take into account motion and contact constraints.

%%%%%%%%%%%%%%%%%%%%%%%%%%%%%%%%%%%%%%%%%%%%%%%%%%%%%%%%%%%%%%%%%%%%%%%%%%%%%%%%

%%%%%%%%%%%%%%%%%%%%%%%%%%%%%%%%%%%%%%%%%%%%%%%%%%%%%%%%%%%%%%%%%%%%%%%%%%%%%%%%

%%%%%%%%%%%%%%%%%%%%%%%%%%%%%%%%%%%%%%%%%%%%%%%%%%%%%%%%%%%%%%%%%%%%%%%%%%%%%%%%

% \section*{ACKNOWLEDGMENT}

%%%%%%%%%%%%%%%%%%%%%%%%%%%%%%%%%%%%%%%%%%%%%%%%%%%%%%%%%%%%%%%%%%%%%%%%%%%%%%%%

\newpage
\bibliographystyle{ieeetr}
\bibliography{references.bib}
% \bibliography{refs_2}

\end{document}